\documentclass[10pt,twocolumn,letterpaper]{article}

\usepackage[pagenumbers]{wacv} 

\usepackage{graphicx}
\usepackage{amsmath}
\usepackage{amssymb}
\usepackage{booktabs}
\usepackage{xcolor}

\usepackage{enumitem}
\usepackage{multirow}
\usepackage{wrapfig}
\usepackage{colortbl}
\usepackage[normalem]{ulem}
\usepackage{microtype}
\usepackage{comment}
\usepackage[hang,flushmargin]{footmisc}

\setlist[itemize]{align=parleft,left=0pt,topsep=1mm,itemsep=0mm,parsep=1mm}

\definecolor{azure(colorwheel)}{rgb}{0.0, 0.5, 1.0}
\definecolor{nicegreen}{rgb}{0.0, 0.7, 0.1}
\definecolor{yw}{rgb}{0.01176, 0.5490, 0.5490}
\definecolor{ashblue}{rgb}{0.36, 0.54, 0.66}
\definecolor{ashgrey}{rgb}{0.7, 0.75, 0.71}
\definecolor{applegreen}{rgb}{0.55, 0.71, 0.0}
\definecolor{blue}{rgb}{0.0, 0.0, 1.0}
\definecolor{postechred}{rgb}{0.784, 0.003, 0.313}
\definecolor{ywg}{rgb}{0.9960, 0.8984, 0.5859}
\definecolor{ballblue}{rgb}{0.13, 0.67, 0.8}
\definecolor{cornellred}{rgb}{0.7, 0.11, 0.11}
\definecolor{darkcyan}{rgb}{0.0, 0.55, 0.55}
\definecolor{CuGray}{gray}{0.9}
\definecolor{airforceblue}{rgb}{0.36, 0.54, 0.66}
\definecolor{rev}{rgb}{0.784, 0.003, 0.313}
\definecolor{pink}{cmyk}{0, 0.7808, 0.4429, 0.1412}
\definecolor{amethyst}{rgb}{0.6, 0.4, 0.8}
\definecolor{black}{rgb}{0.0, 0.0, 0.0}
\definecolor{tb3_yellow}{rgb}{0.996, 1.0, 0.6}
\definecolor{tb3_orange}{rgb}{0.980, 0.8, 0.604}
\definecolor{tb3_red}{rgb}{0.972, 0.6, 0.6}
\definecolor{dimgray}{rgb}{0.41, 0.41, 0.41}
\definecolor{brickred}{rgb}{0.8, 0.25, 0.33}
\definecolor{bleudefrance}{rgb}{0.19, 0.55, 0.91}
\definecolor{blue(ncs)}{rgb}{0.265, 0.445, 0.765}
\definecolor{blue(ryb)}{rgb}{0.01, 0.28, 1.0}
\definecolor{orange}{rgb}{1.0, 0.49, 0.0}
\definecolor{Gray}{gray}{0.88}
\definecolor{green(ncs)}{rgb}{0.0, 0.62, 0.42}
\definecolor{brightpink}{rgb}{1.0, 0.0, 0.5}

\definecolor{kellygreen}{rgb}{0.3, 0.73, 0.09}

\newcolumntype{g}{>{\columncolor{CuGray}}c}
\newcolumntype{z}{>{\columncolor{CuGray}}l}

\renewcommand{\paragraph}[1]{\vspace{1mm}\noindent\textbf{#1.}\,\,}

\usepackage{xspace}

\makeatletter
\def\@fnsymbol#1{\ensuremath{\ifcase#1\or *\or \dagger\or \ddagger\or
   \mathsection\or \mathparagraph\or \|\or **\or \dagger\dagger
   \or \ddagger\ddagger \else\@ctrerr\fi}}
\makeatother

\def\onedot{.\@\xspace}
\def\eg{\emph{e.g}\onedot} 
\def\ie{\emph{i.e}\onedot}

\newcommand{\Sref}[1]{Sec.~\ref{#1}}
\newcommand{\Eref}[1]{Eq.~(\ref{#1})}
\newcommand{\Fref}[1]{Fig.~\ref{#1}}
\newcommand{\Tref}[1]{Table~\ref{#1}}

\newcommand{\be}{\begin{eqnarray}}
\newcommand{\ee}{\end{eqnarray}}
\newcommand{\bee}{\begin{eqnarray*}}
\newcommand{\eee}{\end{eqnarray*}}

\newcommand{\matrixb}{\left[ \begin{array}}
\newcommand{\matrixe}{\end{array} \right]}

\usepackage{mathtools}

\DeclarePairedDelimiter{\norm}{\lVert}{\rVert}

\usepackage[pagebackref,breaklinks,colorlinks]{hyperref}

\usepackage[capitalize]{cleveref}
\crefname{section}{Sec.}{Secs.}
\Crefname{section}{Section}{Sections}
\Crefname{table}{Table}{Tables}
\crefname{table}{Tab.}{Tabs.}

\def\wacvPaperID{2538}
\def\confName{WACV}
\def\confYear{2024}

\begin{document}

\title{LaughTalk: Expressive 3D Talking Head Generation with Laughter}

\def\authorBlock{
    Kim Sung-Bin${}^{1}$ \qquad
    Lee Hyun${}^{1}$ \qquad
    Da Hye Hong${}^{2}$ \qquad \\
    Suekyeong Nam${}^{3}$ \qquad
    Janghoon Ju${}^{3}$ \qquad
    Tae-Hyun Oh${}^{1,4,5}$ \vspace{3mm} \\ 
    ${}^{1}$Dept.~of Electrical Engineering and ${}^{4}$Grad.~School of Artificial Intelligence, POSTECH\\
    ${}^{2}$Sookmyung Women's University \qquad ${}^{3}$KRAFTON\\ 
${}^{5}$Institute for Convergence Research and Education in Advanced Technology, Yonsei University\\
{\normalsize\url{https://laughtalk.github.io/}}\vspace{-2mm}
}
\author{\authorBlock}
\maketitle

\begin{abstract}
   Laughter is a unique expression, essential to affirmative social interactions of humans. Although current 3D talking head generation methods produce convincing verbal articulations, they often fail to capture the vitality and subtleties of laughter and smiles despite their importance in social context. In this paper, we introduce a novel task to generate 3D talking heads capable of both articulate speech and authentic laughter. Our newly curated dataset comprises 2D laughing videos paired with pseudo-annotated and human-validated 3D FLAME parameters and vertices. Given our proposed dataset, we present a strong baseline with a two-stage training scheme: the model first learns to talk and then acquires the ability to express laughter. Extensive experiments demonstrate that our method performs favorably compared to existing approaches in both talking head generation and expressing laughter signals. We further explore potential applications on top of our proposed method for rigging realistic avatars.
\end{abstract}

\section{Introduction}\label{sec:intro}
Speech-driven 3D facial animation has garnered increasing attention from both academic researchers and industries due to its practical applications. This field offers promising applications in content creation, including computer gaming, film production~\cite{liu2009analysis}, and immersive interactions between humans and machines in virtual realities~\cite{wohlgenannt2020virtual}.
Following the practicality, recent advancements in deep learning techniques
{\let\thefootnote\relax\footnote{\textbf{Acknowledgment.} This work was supported by IITP grant funded by Korea government (MSIT) (No.2021-0-02068, Artificial Intelligence Innovation Hub; No.RS-2023-00225630, Development of Artificial Intelligence for Text-based 3D Movie Generation; No.2022-0-00290, Visual Intelligence for Space-Time Understanding and Generation
based on Multi-layered Visual Common Sense; No.2022-0-00124, Development of Artificial Intelligence Technology for Self-Improving Competency-Aware Learning Capabilities).
}}
have yielded impressive outcomes in generating high-quality animated speech-driven 3D talking heads~\cite{voca, karras2017audio, meshtalk, faceformer, codetalker, imitator}.

\begin{figure}[tp]
    \centering
    \includegraphics[width=\linewidth]{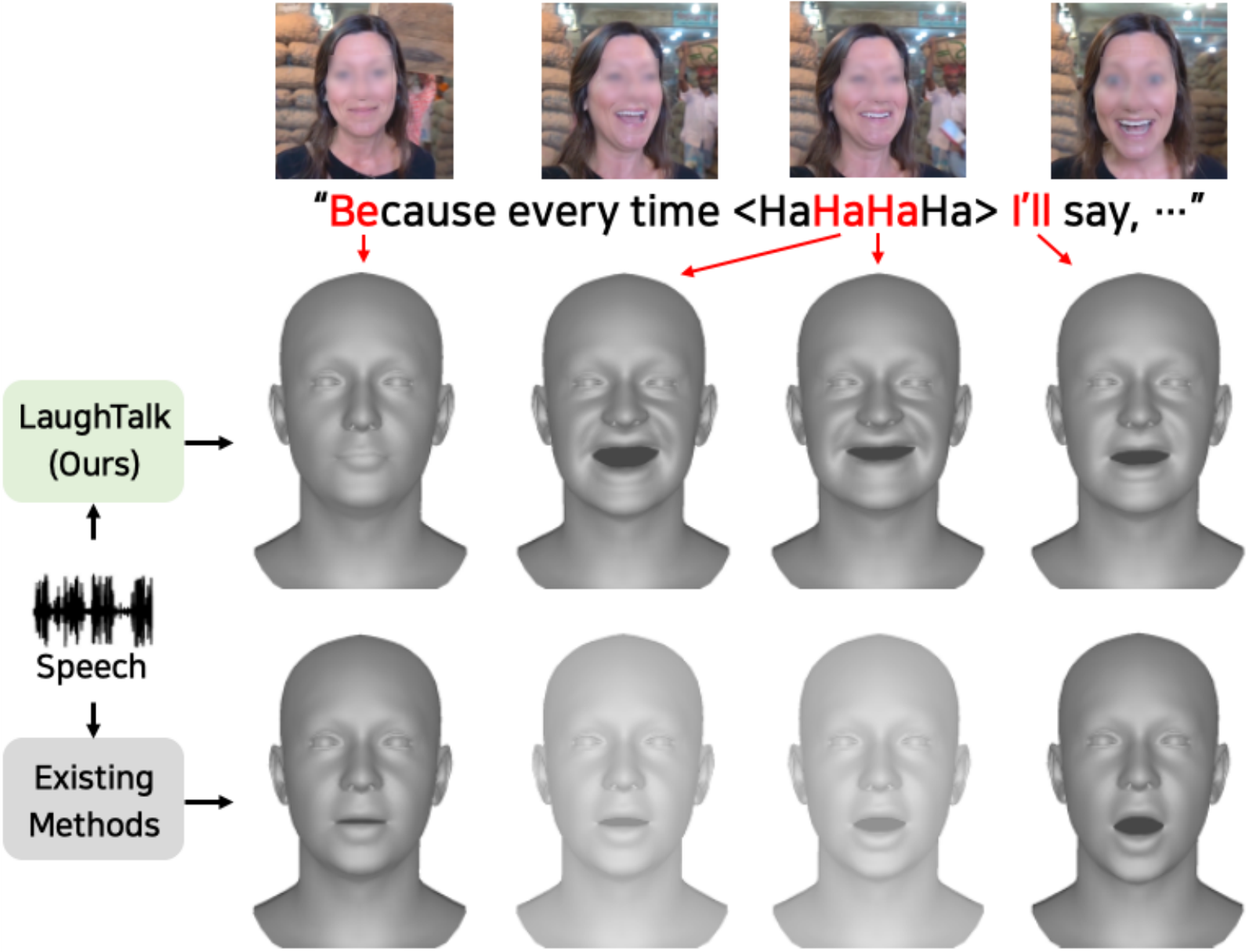}
    \caption{\textbf{Learning to laugh and talk.} We present a task for generating speech-driven 3D talking heads capable of conveying both verbal and non-verbal expressions. In contrast to existing methods that primarily focus on achieving accurate lip synchronization to verbal signals, our goal is to simultaneously articulate synchronized lip movements while also expressing synchronized laughter.}
    \label{fig:teaser}
\end{figure}

However, these prior arts predominantly focus on achieving accurate lip synchronization to verbal signals in speech, often neglecting non-verbal signals, which play vital roles not only in human interactions~\cite{sauter2010perceptual, hinde1972non, mehrabian2017nonverbal} but also in human-computer interaction~\cite{lee2007nonverbal}. 
Among the non-verbal cues, laughter, in particular, holds significance as it is evoked to establish intimacy~\cite{stauffer1999let}, rapport~\cite{adelsward1989laughter}, trust~\cite{vartabedian1993humor}, and create deep emotional exchanges~\cite{scott2014social, tickle1990nature}.
Furthermore, it serves as a powerful medium for conveying diverse social and emotional nuances beyond verbal expressions.
Thus, learning to synthesize speech-driven 3D facial animations incorporating non-verbal expressions, including laughter, may open up the development of intimate and empathetic machines capable of engaging in human-robot interactions.

Nevertheless, accomplishing the animation of 3D facial expressions for both speech and laughter presents intricate challenges. Firstly, the scarcity of comprehensive datasets encompassing both speech and laughter poses a hurdle. Existing 3D scan and speech-paired datasets, such as VOCASET~\cite{voca} and BIWI~\cite{biwi}, are derived from controlled lab environments, lacking the diversity and naturalness required for real-world scenarios. Secondly, verbal and non-verbal cues tend to be entangled within speech. 
This intermingling complicates the extraction of explicit verbal and non-verbal cues from speech, thus making it challenging to teach the 3D talking head to express them simultaneously.
Addressing these challenges is pivotal for solving the 3D facial animation encompassing both speech and laughter.

In this work, we introduce a novel task to animate a 3D face from speech, conveying both talking and laughter. Our primary focus is learning the model to simultaneously articulate the synchronized accurate lip movements and synchronized laughter (\Fref{fig:teaser}). 
For this task, we collect the \textbf{LaughTalk dataset}, comprising in-the-wild 2D videos that feature diverse and natural speech along with laughter, and their corresponding 3DMM (FLAME)~\cite{flame} parameters. Despite the abundance of extensive 2D video datasets~\cite{yu2023celebv, zhu2022celebv, Chung18b}, parsing videos containing both speech and laughter remains non-trivial. We devise a data curation pipeline capable of parsing laughing and talking videos while filtering out noisy ones, such as videos with non-active speakers, scene changes, and abrupt head movements. Since these 2D videos do not contain 3D information, we leverage 3D face reconstruction models~\cite{emoca, spectre} and exemplar fine-tuning methods~\cite{eft, youwang2023largescale} to generate reliable and robust pseudo ground truth FLAME parameters corresponding to the collected 2D videos.

Given this dataset, we design a baseline model for our task, called~\textbf{LaughTalk}. LaughTalk adopts a two-stage training procedure, initially learning speech articulation and subsequently learning to express laughter. The model trained in the first stage (stage-1) entails extracting audio features from a pre-trained audio encoder~\cite{wav2vec} and training a Transformer~\cite{transformer} decoder to regress the FLAME parameters from the audio features. 
This training employs the
subset from the LaughTalk dataset, LaughTalk$_{\text{MEAD}}$\footnote{The video clips are curated from MEAD~\cite{mead}.},
which comprises neutral speech and facial movements, to focus solely on verbal cues. In the second stage, a separate model is trained 
to regress the residual FLAME parameters that the first stage model has not learned. 
The separate model used in the second stage (stage-2) has an identical architecture to the model used in the first stage, but its weights are not shared.
Here, we employ another subset from the LaughTalk dataset, LaughTalk$_{\text{CELEB}}$\footnote{The video clips are curated from CelebV-HQ~\cite{zhu2022celebv} and CelebV-Text~\cite{yu2023celebv}.},
containing both speech and laughter. 
Given that the stage-1 model has already learned to convey speech articulations, the stage-2 model focuses on learning to express non-verbal signals. Combining the FLAME parameters derived from both stage models facilitates the simultaneous representation of verbal and non-verbal cues within the 3D facial animation.

We validate the efficacy of LaughTalk by comparing it with existing 3D talking head models~\cite{voca, meshtalk, faceformer, codetalker}. 
For a fair comparison, we train existing models with our proposed dataset, also allowing them to learn to talk and laugh.
We leverage a pre-trained emotion feature extractor~\cite{mollahosseini2017affectnet} to assess whether the model conveys synchronized non-verbal cues and measure the lip vertex error to evaluate speech articulation. Our experiments demonstrate that LaughTalk not only excels in generating synced laughter but also exhibits favorable lip articulation compared to existing methods. 
Moreover, we showcase the practical application of LaughTalk, further underscoring its potential in real-world scenarios.
Our main contributions are summarized as follows:
\begin{itemize}
    \item Introducing a task to generate a 3D talking head that simultaneously expresses speech articulation and laughter.
    \item Collecting and curating the LaughTalk dataset, which includes 2D videos of speech and laughter along with their corresponding pseudo ground truth FLAME parameters. 
    \item Proposing a baseline, LaughTalk, and a two-stage training scheme for learning to animate expressive 3D faces with verbal and non-verbal signals from speech. 
\end{itemize}
%-------------------------------------------------------------------------
\section{Related work}
\paragraph{Speech-driven 3D facial animation}
There has been a lot of research into generating 3D talking heads for 3D gaming and virtual reality applications. Seminal works~\cite{cao2005expressive, karras2017audio, tian2019audio2face, voca, hussen2020modality, faceformer, meshtalk, emote, emotalk} have shown promising results in speech-driven 3D facial animation, particularly in generating verbal articulations in sync with input speech. 
FaceFormer~\cite{faceformer} uses a Transformer-based model for the first time to autoregressively generate facial movements based on speech input. 
CodeTalker~\cite{codetalker} learns a discrete codebook for generic facial motion and leverages a similar architecture from FaceFormer for animation synthesis. 
However, they mainly focus on the mouth movement, which is related to the verbal context of speech. 
For enhancing facial expression, Meshtalk~\cite{meshtalk} aims to capture upper face movements by designing separate latent codes for audio-related and non-audio-related movements, such as blinking and eyebrow raises. 
The aforementioned methods mainly focus on the verbal feature of speech or speech-uncorrelated movements but overlook addressing non-verbal expression arising from speech, which often carries the emotional and social context. 

In more recent works, EMOTE~\cite{emote} and EmoTalk~\cite{emotalk} introduce emotional 3D talking heads. EMOTE requires explicit condition inputs to the model, such as emotion categories (\eg, angry, surprise, sad, etc.). EmoTalk alleviates this explicit conditioning by deriving emotion from speech but requires an artist-curated dataset for training. While these models can generate verbal articulation incorporated with emotion, they cannot capture non-verbal signals, such as laughter, from speech.
In contrast, we focus on generating an expressive 3D talking head that embraces both non-verbal (\ie, laughter) and verbal signals deriving directly from the speech and provide a new dataset suitable for this task. 

\paragraph{Face video datasets}
One of the challenges in speech-driven 3D facial animation is the lack of 3D paired datasets. BIWI~\cite{biwi}, VOCASET~\cite{voca}, S3DFM~\cite{S3DFM}, and Multiface~\cite{multiface} are publicly available datasets for this task. 
Although these datasets provide accurate 3D meshes scanned in lab environments, they are limited in size, diversity, speaking styles, and naturalness. 
On the other hand, many 2D face video datasets~\cite{Nagrani17, Chung18b, Chung17, afouras2018deep, jiang2020deeperforensics, zhu2022celebv, yu2023celebv} are available. 
Most of them are curated from in-the-wild video sources, \eg, YouTube, and have richness in size, diversity, and naturalness. 

Our newly curated dataset is derived from existing 2D face video datasets.
We annotate these videos with reliable pseudo 3DMM parameters, by leveraging the techniques from 3D face reconstruction methods~\cite{emoca, spectre, deca}. 
We believe that utilizing extensive 2D facial video datasets with 3D reconstruction methods is advantageous for building a rich dataset that can be used for training 3D talking heads conveying a wide range of verbal and non-verbal expressions.

\paragraph{Laughter in non-verbal signals}
Among diverse non-verbal signals, we especially focus on laughter, which plays a key role in social interactions, such as building rapport, expressing emotions, and creating deep emotional exchanges ~\cite{scott2014social, tickle1990nature}.
Furthermore, laughter is a distinct human expression not found in other animals, which we often use to establish intimacy~\cite{stauffer1999let}, grab attention~\cite{wanzer2010explanation}, or build trust~\cite{vartabedian1993humor}. 
Understanding and synthesizing laughter is thus a crucial stepping stone toward creating expressive and intimate interactive agents that go beyond verbal-only communication.
\section{Learning to laugh and talk}
In this section, we provide a concise overview and preliminary for our proposed task (\Sref{sec3_1}) and introduce the LaughTalk dataset comprising paired 2D video and 3DMM parameters (\Sref{sec3_2}). Subsequently, we propose LaughTalk, a two-stage training baseline model capable of both laughing and talking synchronized to the given speech (\Sref{sec3_3}).

\subsection{Overview}\label{sec3_1}
Our goal is to synthesize a sequence of 3D face animations from given speech audio, encompassing both speech and laughter. In contrast to the explicit conditioning on global emotion labels, \eg, \cite{emote}, we drive the lips and expressions of 3D faces synchronized to the talking (verbal) and instantaneous laughing (non-verbal) cues in the speech. 
For instance, when the speech is delivered with a neutral tone, the 3D faces focus solely on animating the lips with neutral expressions. Conversely, during laughing in the speech, the 3D facial features would synchronously animate, depicting the characteristic upward shift of expression associated with laughter. 
As we take the pioneering step, we aim to build an appropriate 3D face dataset and baseline model for this task.
\begin{figure}[t]
  \centering
\includegraphics[width=1\linewidth]{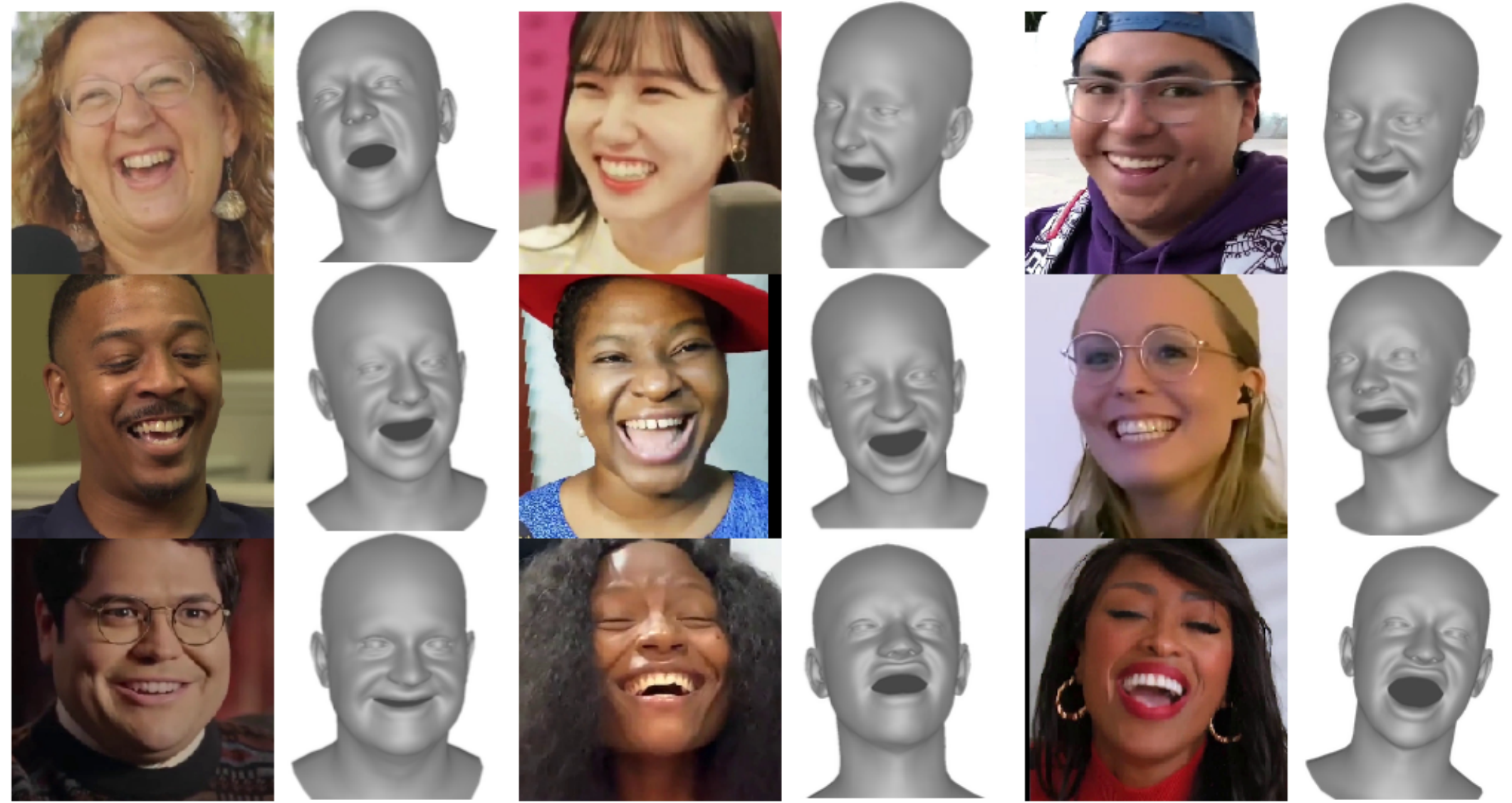}
   \caption{\textbf{LaughTalk Dataset.} We collect and curate 2D talking and laughing videos with corresponding pseudo 3DMM parameters. Here, we show the images sampled from the 2D videos and corresponding mesh images rendered from the 3DMM parameters.
   }
   \label{fig:data}
\end{figure}

\paragraph{Preliminary}
We use FLAME~\cite{flame}, a parametric 3D head model, as the 3D human face representation for our collected dataset and the 3D talking head.
Given the parameters of the face shape $\boldsymbol{\beta}\in{\mathbb R}^{|\boldsymbol{\beta}|}$, facial expression coefficients $\boldsymbol{\psi}\in{\mathbb R}^{|\boldsymbol{\psi}|}$, and pose $\boldsymbol{\theta}\in{\mathbb R}^{3k+3}$ ($k{=}4$ joints), 3D face mesh with vertices $\mathbf{V}\in{\mathbb R}^{n_v\times 3}$ ($n_v{=}5023$) and 3D facial landmarks $\mathbf{J}^{\text{3D}}\in{\mathbb R}^{n_j\times 3}$ ($n_j{=}68$) can be acquired with a differentiable FLAME model $\boldsymbol{M}$, where
$[\mathbf{V}, \mathbf{J}^{\text{3D}}] = \boldsymbol{M}(\boldsymbol{\beta}, \boldsymbol{\psi}, \boldsymbol{\theta})$.

\subsection{LaughTalk dataset}\label{sec3_2}
We introduce the LaughTalk dataset, comprising talking and laughing facial video clips with corresponding pseudo-annotated FLAME parameters. 
We first curate 2D video clips from MEAD~\cite{mead} for neutral talking (called LaughTalk$_{\text{MEAD}}$), and CelebV-HQ~\cite{zhu2022celebv} and CelebV-Text~\cite{yu2023celebv} for capturing talking with laughter (called LaughTalk$_{\text{CELEB}}$). 
Then, we filter out the clips to select valid samples and finally obtain 3D pseudo-annotation from them.
Examples of our curated dataset are shown in Fig.~\ref{fig:data}.

\paragraph{Data collection}
CelebV-HQ and CelebV-Text have in-the-wild face videos with rich facial attributes, such as appearance, emotion, and action. Among the attributes, we query ``laugh'', ``smile'', ``happy'', and ``talk'' to construct laughing and talking video clips for LaughTalk$_{\text{CELEB}}$. 
Similarly, MEAD provides the annotations of emotion attributes and their intensities.
We query the ``neutral'' attribute in MEAD and collect neutral talking video clips for LaughTalk$_{\text{MEAD}}$.
\begin{table}
  \centering
  \resizebox{0.93\linewidth}{!}{
    \begin{tabular}{@{\quad}l@{\quad\quad\quad} c@{\quad}}
    \toprule
    \multicolumn{2}{c}{Statistics of our LaughTalk dataset}\\
    \midrule
    Number of Total Video Clips & 943 \\
    Number of LaughTalk$_{\text{MEAD}}$ & 438 \\
    Number of LaughTalk$_{\text{CELEB}}$ & 505 \\
    Number of Train / Test of LaughTalk$_{\text{MEAD}}$&  374 / 64  \\
    Number of Train / Test of LaughTalk$_{\text{CELEB}}$ &  455 / 50  \\
    Average Duration of Train set & 3.5 sec. \\
    Average Duration of Test set & 5.6 sec. \\
    \bottomrule
    \end{tabular}  
    }
    \vspace{-2mm}
    \caption{\textbf{Statistics of the LaughTalk dataset.} We trim training set video clips to 3.5 seconds but leave the test set with varying lengths for evaluation on diverse inputs. 
    Thus, the average length of the test set video is longer than the training set.} 
    \vspace{-4mm}
  \label{tab:dataset_statistics}
\end{table}

\paragraph{Data filtering process}
After the data collection, we filter out noisy samples to construct a valid and clean dataset. 
First, to ensure that video clips always contain laughter, we use a laugh detector~\cite{gillick2021robust} to filter out samples that do not have laughter for at least 3.5 seconds. 
This reduces false samples from incorrect attribute annotations in the original datasets. 
Second, our dataset must include talking faces. However, some video clips contain speech from outside the scene. 
Thus, we filter out the videos with non-active speakers using the active speaker detector~\cite{tao2021someone}, ensuring our dataset holds the facial video that synchronizes with speech.
Third, we use the scene detector~\cite{PySceneDetect} to trim video clips at scene transitions, thereby preventing the inclusion of scene change videos. 
Only videos longer than 3.5 seconds are considered in this process. 
Lastly, we exclude video samples in which the individual's face is not visible from the front, only a partial view of the face is shown, or there are abrupt head movements. 
We retain only clear, frontal facial shots.

\paragraph{Lifting 2D video to 3D}
After acquiring the cleaned 2D in-the-wild videos, we reconstruct 3D faces synchronized with both the audio and facial movements from the video clips. 
However, existing 3D face reconstruction models~\cite{deca, emoca} have limitations for reconstructing temporally consistent and accurate 3D face meshes from videos. 
State-of-the-art face reconstruction models are typically trained exclusively on static 2D images. 
This results in limitations when extrapolating to faces in rare poses and produces jittered motion due to per-frame independent inference.
To address these challenges, we employ an optimization method~\cite{youwang2023largescale} that re-parameterizes 3D face meshes with neural network parameters, inspired by EFT~\cite{eft}.
We initialize the neural network with SPECTRE~\cite{spectre} and optimize it for each video clip, ensuring the acquisition of accurate and robust pseudo ground truth FLAME parameters. 
This approach is suitable for our purpose as it yields an accurate 3D face reconstruction result that best fits each video clip.

After all the processing, we have 943 video clips and corresponding pseudo-annotated FLAME parameters. We set separate training and test sets for each sub-dataset (LaughTalk$_{\text{MEAD}}$, LaughTalk$_{\text{CELEB}}$).
The basic statistics for our dataset are summarized in Table~\ref{tab:dataset_statistics}.

\subsection{Two-stage training baseline: LaughTalk}\label{sec3_3}
\begin{figure*}[th]
  \centering
\includegraphics[width=1\linewidth]{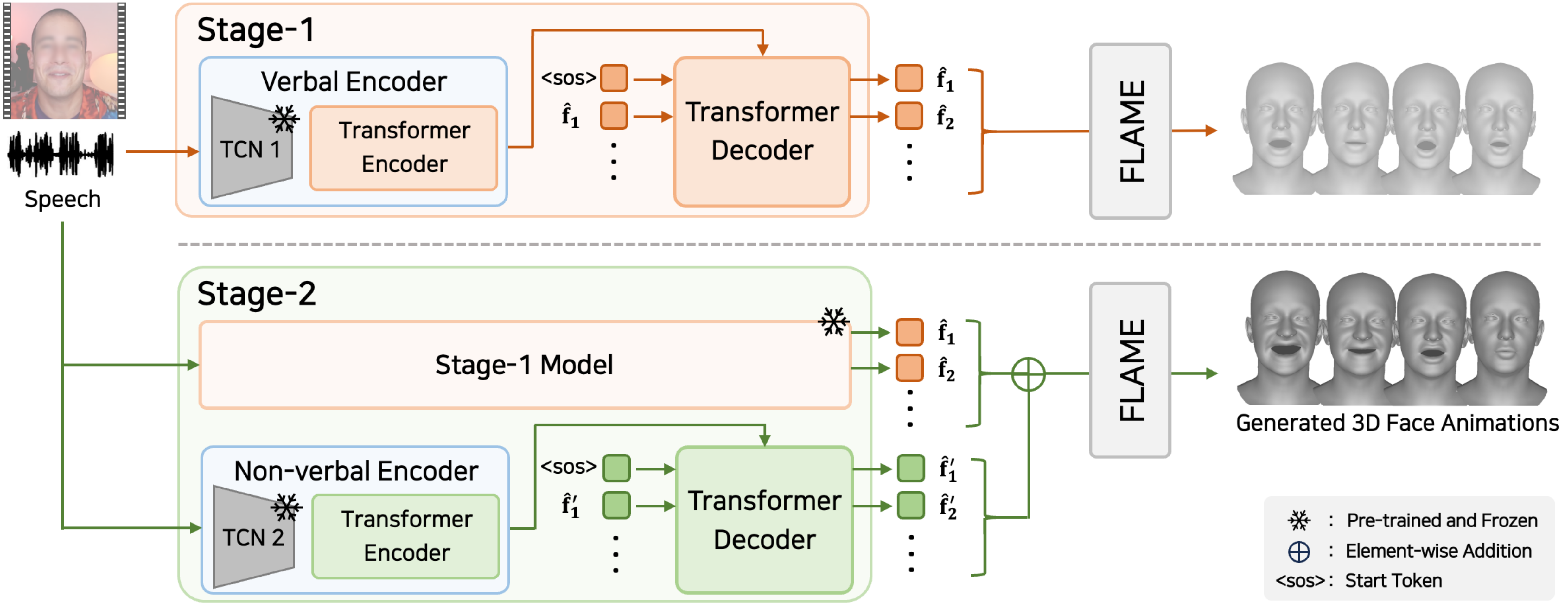}
   \caption{\textbf{LaughTalk architecture.}
   The stage-1 model extracts verbal features from the input speech and generates facial motion representations in an autoregressive manner. Simultaneously, the stage-2 model extracts non-verbal features from the same speech and generates residual facial motion representations. These two sets of representations are element-wise added and subsequently fed into the FLAME model to synthesize 3D face animations.
   }
   \label{fig:system}
\end{figure*}

Since verbal and non-verbal signals are often intertwined within a single speech, teaching a 3D talking head model to animate both laughter and speech simultaneously is challenging. To address this, we approach the task by breaking it down into sub-problems. The schematic overview of our proposed baseline model, LaughTalk, is presented in \Fref{fig:system}. 

LaughTalk undergoes a two-stage training strategy. 
First, the stage-1 model learns to talk (\ie, verbal signals from speech). 
It focuses on learning to generate facial representations for lip movement synchronization with neutral speech videos (LaughTalk$_{\text{MEAD}}$). 
After the stage-1 model has acquired the ability to animate speech-related movements, we then progress to train the stage-2 model using LaughTalk${_\text{CELEB}}$, aiming to animate both lip movements and facial expressions simultaneously. 
As we freeze the parameters of the stage-1 model, the stage-2 model focuses on learning to generate residual facial representations that are not learned by the stage-1 model. 
These residual aspects likely correspond to non-verbal cues present in the speech, such as cheek movements and facial expressions. 
By combining the residual representations with the output of the pre-trained stage-1 model, LaughTalk can simultaneously animate a 3D talking head with synchronized verbal and non-verbal signals.
We now describe the brief task formulation, details of our model's architecture, and the training objectives for each stage.

\paragraph{Task formulation}
Let $\mathbf{F}_{1:T}=(\mathbf{f}_1,\dots, \mathbf{f}_T)$ be a temporal sequence of facial motions, with each frame $\mathbf{f}_t$ denoting the facial representation. 
Here, we define the facial representation $\mathbf{f}_t$ as the concatenation of the expression coefficient and jaw pose of the FLAME parameters, 
$\mathbf{f}_t=[\boldsymbol{\psi}_t,\boldsymbol{\theta}^{\text{jaw}}_t]$.
Additionally, let $\mathbf{A}_{1:T}=(\mathbf{a}_1,\dots, \mathbf{a}_T)$ be a sequence of speech representations.
Then, our goal is to sequentially synthesize the facial representations $\mathbf{F}_{1:T}$ from the corresponding $\mathbf{A}_{1:T}$.
To visualize $\mathbf{F}_{1:T}$ as mesh vertices, these facial representations are fed into the FLAME model $\boldsymbol{M}$ along with arbitrary face shape parameter $\boldsymbol{\beta}$, generating animated vertices as $\mathbf{V}_{1:T}=\boldsymbol{M}(\boldsymbol{\beta}, \mathbf{F}_{1:T})$.

\paragraph{Stage-1: learning to talk}
The stage-1 model is designed to learn to animate the 3D faces, primarily capturing speech-related signals.  
This model comprises the Verbal Encoder $E_v$, which extracts speech-related features from the input audio, and the Transformer Decoder $D_v$, which takes the speech-related features and generates sequences of facial representations in an autoregressive manner (Stage-1 in \Fref{fig:system}).

Following Faceformer~\cite{faceformer}, we employ wav2vec 2.0~\cite{wav2vec} for the Verbal Encoder. This includes the audio feature extractor and multi-layer transformer encoder.
The audio feature extractor utilizes a temporal convolutional network (TCN) to convert the raw waveform of speech into feature vectors.
Then, the Transformer Encoder transforms the audio features into speech representations. 
The Transformer Decoder is equipped with causal self-attention to learn the dependencies within the context of previous facial representations and employs cross-modal attention to align the audio and facial representations.
Formally, this process can be written as: $\mathbf{\hat{f}}_t=D_v(E_v(\mathbf{A}_{1:T}),\mathbf{\hat{F}}_{1:t-1})$, where $\mathbf{\hat{f}}_t$ is the currently predicted facial representation, and $\mathbf{\hat{F}}_{1:t-1}$ is the past predicted sequences. 
After predicting all the sequences $\mathbf{\hat{F}}_{1:T}$, we feed these along with arbitrary shape parameter $\boldsymbol{\beta}$ to the FLAME model $\boldsymbol{M}$ to convert them to the mesh vertices and 3D landmarks, 
$[\mathbf{\hat{V}}_{1:T}, \mathbf{\hat{J}^{\text{3D}}}_{1:T}] = \boldsymbol{M}(\boldsymbol{\beta}, \mathbf{\hat{F}}_{1:T})$.

We train the Transformer Encoder and Decoder of the stage-1 model while keeping TCN frozen with the pre-trained weights of wav2vec 2.0. 
To ensure that the model learns to animate the facial movements by mainly focusing on the verbal signals, we train it using LaughTalk$_{\text{MEAD}}$, which contains speech delivered with neutral tones. The training objective for the stage-1 model is:
\begin{equation}\label{loss1}
    \textstyle L_{\text{stage-1}} = \lambda_{\text{exp}}\norm{\boldsymbol{\psi}_{1:T}-\boldsymbol{\hat{\psi}}_{1:T}}_2+\lambda_{\text{lmk}}\norm{\mathbf{J}^{\text{3D}}_{1:T}-\mathbf{\hat{J}}^{\text{3D}}_{1:T}}_2,
\end{equation}
where $\textstyle \boldsymbol{\hat{\psi}}_{1:T}$ is the expression parameters from the predicted facial representations $\textstyle \mathbf{\hat{F}}_{1:T}$, $\textstyle \mathbf{\hat{J}}^{\text{3D}}_{1:T}$ is the predicted 3D facial landmarks, and $\textstyle \{\lambda_*\}$ denotes the weights for each loss term.

\paragraph{Stage-2: learning to laugh}
The stage-2 model incorporates the pre-trained stage-1 model and is designed to learn to animate 3D faces with both speech and laughter. 
As we freeze the stage-1 model, the stage-2 model is induced to focus on learning to generate residual facial representations that the stage-1 model has not learned. 
We postulate that such residual representations likely correspond to non-verbal cues within speech. 
In other words, the stage-2 model learns to predict the residual facial representations $\mathbf{\hat{F}}^{'}_{1:T}$, which would make the 3D face express both verbal and non-verbal signals, when combined with the outputs of the stage-1 model $\mathbf{\hat{F}}_{1:T}$ (Stage-2 in \Fref{fig:system}).

The architecture of the stage-2 model is identical to that of the stage-1, comprising the Non-verbal Encoder $E_n$ and the Transformer Decoder $D_n$.
Different from stage-1, the stage-2 model uses wav2vec 2.0 for the Non-verbal Encoder, initialized with the pre-trained weights of the emotion recognition model\footnote{\url{https://huggingface.co/harshit345/xlsr-wav2vec-speech-emotion-recognition.}}.

We train the Transformer Encoder and Decoder on LaughTalk$_{\text{CELEB}}$, which contains in-the-wild audio samples with both laughter and speech, while keeping TCN and the stage-1 model frozen. 
The training objective for the stage-2 model is:
\begin{equation}\label{loss2}
    L_{\text{stage2}} = \norm{\mathbf{V}_{1:T}-\mathbf{\hat{V}}^{'}_{1:T}}_2,
\end{equation}
where $\mathbf{V}_{1:T}$ and $\mathbf{\hat{V}}^{'}_{1:T}$ are the sequence of the ground truth and predicted vertices. The predicted vertices are made by summing the FLAME parameters from the pre-trained stage-1 model and the residual part of the stage-2 model. Formally, we denote $\mathbf{\hat{V}}^{'}_{1:T}$ as $\mathbf{\hat{V}}^{'}_{1:T} = \boldsymbol{M}(\boldsymbol{\beta}, \mathbf{\hat{F}}_{1:T}+\mathbf{\hat{F}}^{'}_{1:T})$.

\begin{figure}[tp]
    \centering
    \includegraphics[width=0.8\linewidth]{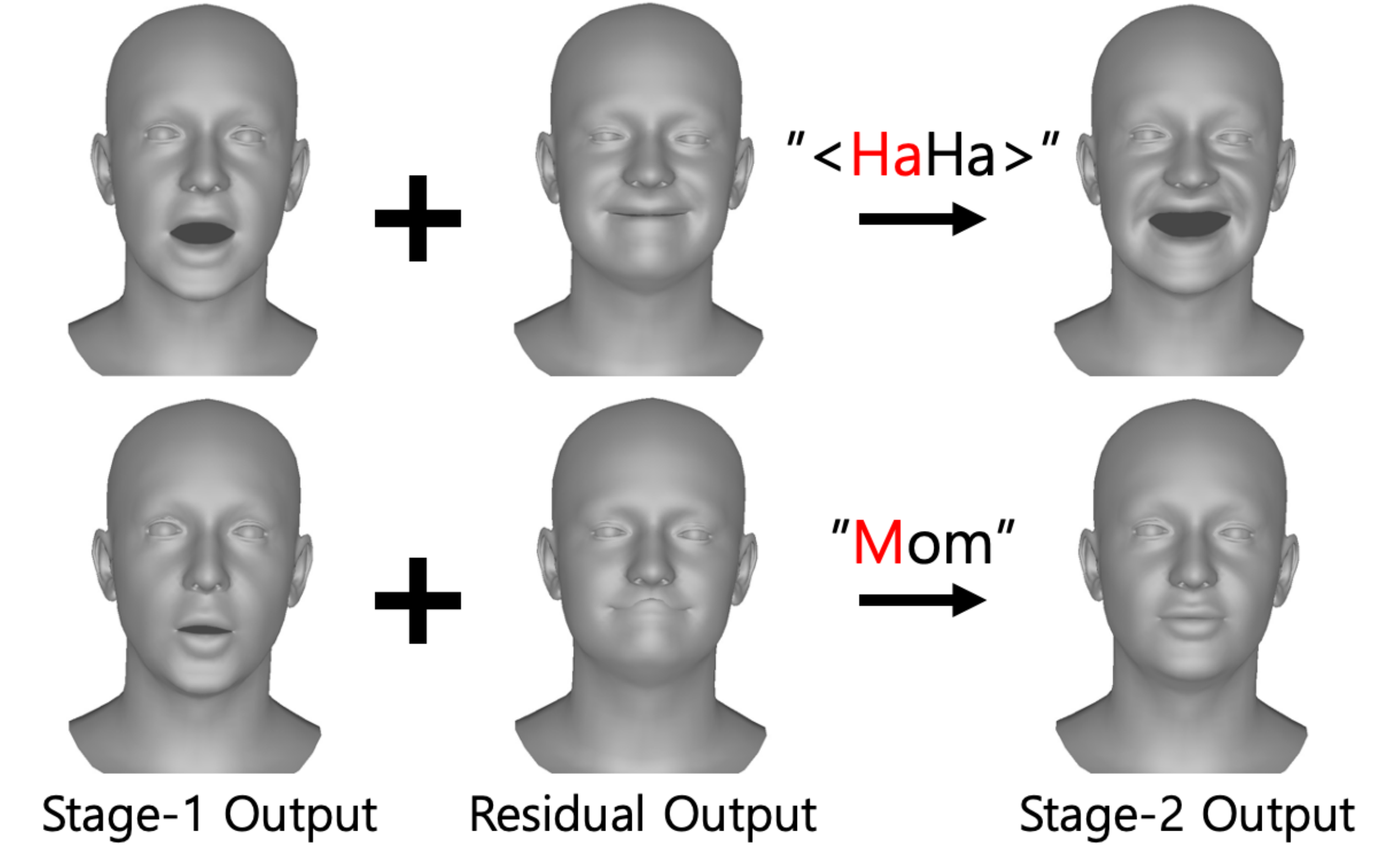}
    \caption{\textbf{Visualization of each model's output.} We visualize the output meshes synthesized by the models from each stage. The stage-1 model outputs the lip movements for the input speech, while the stage-2 model complements the expression of non-verbal signals, and contributes to more accurate lip movements.}
    \label{fig:qual2}
\end{figure}

To dissect the impact of the residual representations learned within the stage-2 model, we convert the facial representation generated by each model into mesh vertices.
Figure~\ref{fig:qual2} visually illustrates the outputs of the stage-1, residual, and the final stage-2 model.
While the outputs of the stage-1 model are geared towards animating speech-related facial motions, the residual outputs contribute to producing more expressive 3D faces. These residual representations serve the dual purpose of conveying the non-verbal laughter signal while also enhancing lip articulations. These findings support the efficacy of the two-stage training approach, successfully animating a 3D talking head with speech and laughter simultaneously.

\paragraph{Training details}
We train the stage-1 model on a single GeForce RTX 3090 for 100 epochs with early stopping. We use the Adam optimizer, and set the batch size to 1 and the learning rate to $2\cdot10^{-4}$. 
After training, we freeze the stage-1 model and train the stage-2 model with the same experiment setup.
For both stages, we randomly sample 100 frames from the input data for training.
\section{Experiments}
We evaluate the performance of LaughTalk using our proposed dataset, as detailed in \Sref{sec3_2}. 
We begin with a quantitative assessment to measure the accuracy of both lip articulation and laughter synchronization in relation to the input speech (\Sref{quan}). We then conduct qualitative comparisons against existing methods (\Sref{qual}). Most importantly, we conduct user studies to validate the human perceptual experience of the generated 3D talking head, encompassing both verbal and non-verbal signals (\Sref{user}). Finally, we show ablation studies to verify our design choices of the model training (\Sref{abla}). 

\subsection{Quantitative evaluation}\label{quan}
To evaluate the lip synchronization to the speech, we measure the lip vertex error (LVE) metric proposed in MeshTalk~\cite{meshtalk}.
The LVE computes the average $\ell_2$ error between the lip regions of the generated mesh vertices and the ground truth from the test set.
For each frame, the LVE is defined as the maximum $\ell_2$ error across all lip vertices\footnote{We use the lip vertex indexes which are provided at \url{https://flame.is.tue.mpg.de/}.}.

However, solely measuring the $\ell_2$ error of lip vertices is insufficient to assess the facial movement synchronization to the laughter, as there is no one-to-one mapping between input audio and the generated 3D faces. 
To address this, we introduce the emotional feature distance (EFD) as an additional metric to assess laughter synchronization.
This metric leverages AffectNet~\cite{mollahosseini2017affectnet}, an emotion recognition model.
Specifically, using AffectNet, we extract the emotion features from both the sequence of rendered images and the frames from ground truth 2D video, then compute the frame-by-frame feature $\ell_2$ distance.
We average these feature distances to determine the EFD.
Our rationale behind this approach is that achieving a lower feature distance may suggest closer temporal and semantic alignments of facial expressions between the generated mesh and the ground truth 2D video, indicating improved synchronization.
\begin{table}[t]
\footnotesize
\centering
    \resizebox{0.97\linewidth}{!}{
    \begin{tabular}{l@{\quad\quad}c@{\quad\quad}c@{\quad\quad}c}
    \toprule
    \multirow{2}{*}{Method}&LaughTalk$_{\text{MEAD}}$&\multicolumn{2}{c}{LaughTalk$_{\text{CELEB}}$}\\
    & LVE ($\downarrow$) & LVE ($\downarrow$) & EFD ($\downarrow$)\\
    \cmidrule{1-4}
    VOCA~\cite{voca} &2.45 & 2.17&17.71\\
    FaceFormer~\cite{faceformer} &\textbf{2.15} & \uline{1.81}&\uline{17.51}\\
    CodeTalker~\cite{codetalker} &2.81 & 2.22&19.80\\
    LaughTalk (Ours) &\uline{2.27} &\textbf{1.76}&\textbf{17.37}\\
    \bottomrule
    \end{tabular}
    }
    \caption{\textbf{Quantitative comparison to existing methods.} We compare LaughTalk (Ours) with existing methods trained on the LaughTalk dataset. The results show that LaughTalk performs favorably in the lip vertex error (LVE), while outperforming other methods in the emotion feature distances (EFD). The measurement scale for LVE is $\times 10^{-4}\text{mm}$ scale. We highlight the best results in \textbf{bold} and \underline{underline} the second best among all the methods.
     }
\label{tab:quan}
\end{table}

We conduct a quantitative comparison of our approach against three state-of-the-art methods, VOCA~\cite{voca}, Faceformer~\cite{faceformer}, and Codetalker~\cite{codetalker}.
To ensure a fair comparison, we retrain these models using the mesh vertices from our LaughTalk dataset.
Table~\ref{tab:quan} summarizes the LVE and EFD over the test set of the LaughTalk dataset.
Notably, LaughTalk performs favorably compared to the other methods across all test sets.
Particularly for the LaughTalk$_{\text{CELEB}}$ test set, encompassing both speech and laughter, our approach demonstrates superior performance in terms of both LVE and EFD metrics. 
This result highlights the effectiveness of our proposed method, LaughTalk, in achieving accurate synchronization of both lip movements and laughter with the corresponding speech.
\begin{figure}[tp]
    \centering
    \includegraphics[width=\linewidth]{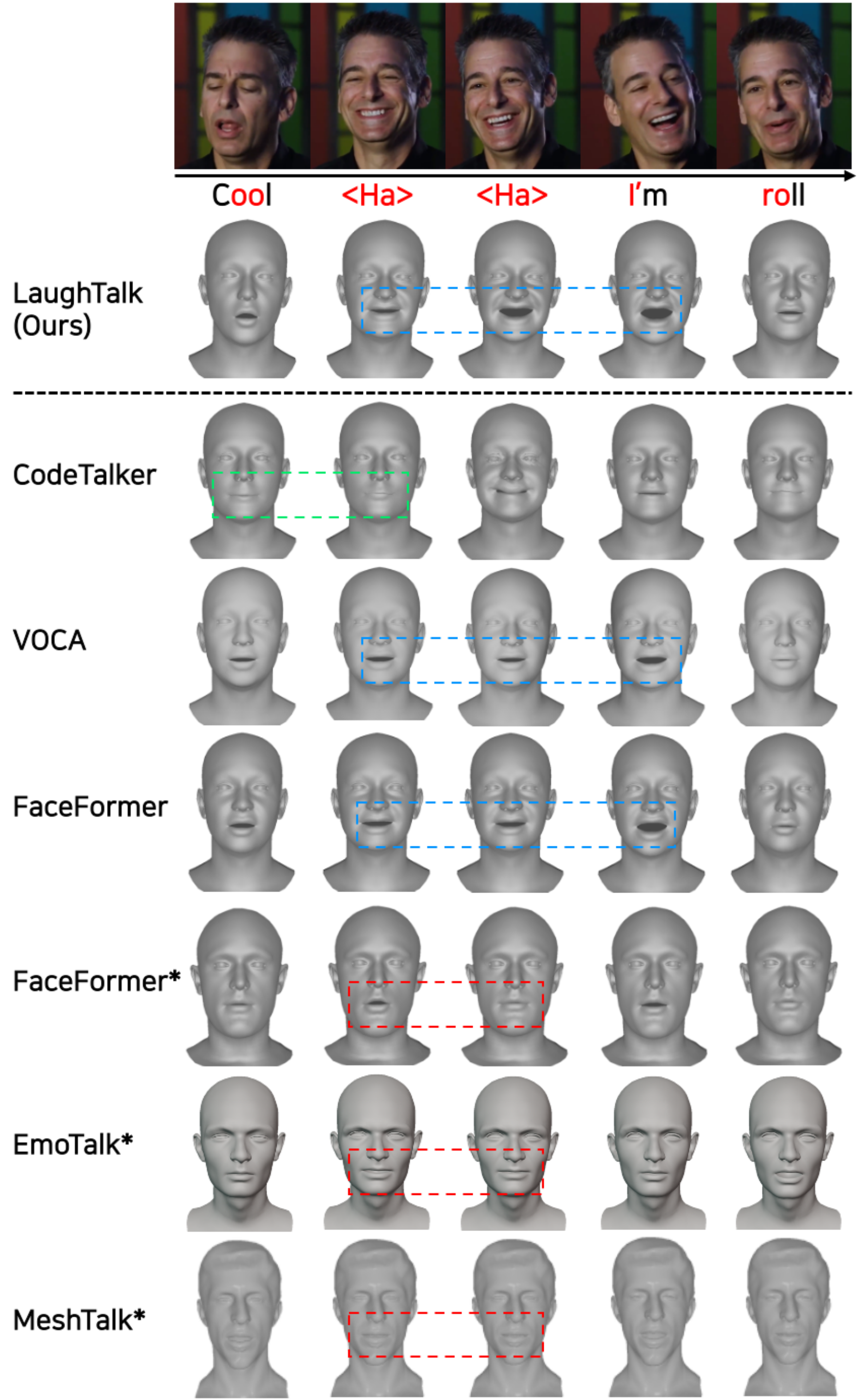}
    \vspace{-7mm}
    \caption{\textbf{Qualitative comparison to existing methods.} Each row shows specific frames synthesized by LaughTalk (Ours) and counterpart methods.
    The models presented in sequence, Ours, CodeTalker, VOCA, and FaceFormer, are trained with our proposed LaughTalk dataset. 
    The models with superscript $^*$, FaceFormer$^*$, EmoTalk$^*$, and MeshTalk$^*$, are the official pre-trained models.}
    \vspace{-4mm}
    \label{fig:qual}
\end{figure}

\subsection{Qualitative evaluation}\label{qual}
Since evaluating speech-driven 3D facial movements based on only quantitative metrics might not capture the full scope of their quality, we visualize the synthesized meshes for more comprehensive evaluation. 
In \Fref{fig:qual}, we present a visual comparison between the meshes generated by our approach and existing methods. 

The top three rows display results from the CodeTalker, VOCA, and FaceFormer models trained on the LaughTalk dataset, where we observe subtle smiles synchronized with the laughter in the speech.
However, CodeTalker struggles to accurately animate lip movements aligned with the verbal signal (green dotted line in \Fref{fig:qual}), while VOCA and FaceFormer exhibit less expressive results compared to our approach (blue dotted line in \Fref{fig:qual}).
The bottom three rows showcase the results from models using their pre-trained weights. 
These primarily exhibit facial movements aligned with the verbal cues, yet neglect the laughter signal, as highlighted by the red dotted lines in \Fref{fig:qual}. 
This demonstrates that our two-stage training method enables the model to produce both verbal and non-verbal signals simultaneously.

\begin{table}[t]
\footnotesize
\centering
    \resizebox{1\linewidth}{!}{
    \begin{tabular}{lcccc}
    \toprule
    Competitors&\multirow{2}{*}{Lip Sync}&\multirow{2}{*}{Laughter Sync} & \multirow{2}{*}{Realism} & \multirow{2}{*}{Intimacy}\\
    (\textit{vs.} B)&  &  & &\\
    \cmidrule{1-5}
    \textit{vs.} FaceFormer &88.00&82.67 &86.00& 84.67\\
    \textit{vs.} CodeTalker &78.00&78.00&82.00 &80.00\\
    \cmidrule{1-5}
    \textit{vs.} FaceFormer* &80.67 &93.33& 81.33 & 84.67 \\
    \textit{vs.} CodeTalker* &67.33 &92.00&79.33&82.00\\
    \textit{vs.} EmoTalk* &72.67 &86.67&66.67&81.33\\
    \bottomrule
    \end{tabular}
    }
    \caption{\textbf{User study results.} We employ A vs. B testing and report the percentage (\%) of responses where A (Ours) is preferred over B. A higher percentage indicates better performance of our method. The user study is conducted to evaluate the generated meshes in four aspects: lip sync, laughter sync, realness, and intimacy. The models with superscript $^*$, FaceFormer$^*$, CodeTalker$^*$, and EmoTalk$^*$, are the official pre-trained models.
     }
\label{tab:user}
\end{table}

\subsection{User study}\label{user}
We provide the user study to evaluate our proposed model, LaughTalk.
Given that the human perception system has evolved to effectively understand subtle facial motions and capture lip articulation, employing user studies stands as a reliable measure for assessing the quality of speech-driven facial animation.
We generate fifteen 3D talking head videos using our method (A) and other methods (B) from the test split of LaughTalk${\text{CELEB}}$ and design a user questionnaire based on A vs.~B testing.
This questionnaire prompts users to choose between two samples based on four distinct aspects: lip synchronization, laughter synchronization, realism, and intimacy. 
Notably, the ``Intimacy'' aspect assesses which generated facial animation elicits a stronger sense of intimacy for human-computer interaction~\cite{lee2007nonverbal}. 
A total of 50 participants take part in this user study.

We compare our model with existing methods trained on the LaughTalk dataset and official pre-trained models.
The user study results are summarized in \Tref{tab:user}, indicating that the participants favor the generated results of LaughTalk over counterpart methods. 
Particularly in the ``Laughter Sync'' and ``Intimacy'' evaluations, our model gets significant preference over the competing methods.
We believe this preference is attributed to two key factors. 
First, our two-stage training design accurately articulates lip movements in the first stage and learns residual features in the second stage to enhance expressiveness and convey non-verbal signals.
Second, our model's ability to achieve accurate lip synchronization while conveying non-verbal signals potentially results in users feeling greater intimacy towards the generated meshes, underscoring the importance of effectively conveying non-verbal signals in 3D talking head models.

\begin{table}[t]
\footnotesize
\centering
    \resizebox{0.95\linewidth}{!}{
    \begin{tabular}{l@{\quad}c@{\quad}c@{\quad}c}
    \toprule
    \multirow{2}{*}{Configurations}&LaughTalk$_{\text{MEAD}}$&\multicolumn{2}{c}{LaughTalk$_{\text{CELEB}}$}\\
    & LVE ($\downarrow$) & LVE ($\downarrow$) & EFD ($\downarrow$)\\
    \cmidrule{1-4}
    w/o LaughTalk$_{\text{CELEBV}}$ & 2.61 & 4.65&20.33\\
    w/o LaughTalk$_{\text{MEAD}}$ & 2.54& 2.24&18.56\\
    \cmidrule{1-4}
    w/o two-stage training &2.43 & 1.86&18.42\\
    \cmidrule{1-4}
    LaughTalk (Ours) &\textbf{2.27} &\textbf{1.76}&\textbf{17.37}\\
    \bottomrule
    \end{tabular}
    }
    \caption{\textbf{Ablation studies for our design choices.} We evaluate on test set of LaughTalk$_{\text{MEAD}}$ and LaughTalk$_{\text{CELEB}}$. While the first two configurations differ in terms of the training set, the subsequent configuration serves as an ablation for the two-stage training strategy. The measurement scale for LVE is $\times 10^{-4}\text{mm}$ scale.}
\label{tab:ablation}
\end{table}

\subsection{Ablation study}\label{abla}
We conduct a series of ablation studies to validate our design choices and assess the effectiveness of our proposed dataset, as summarized in \Tref{tab:ablation}.
Note that, except for our final LaughTalk model, all the ablation models are trained only in stage-1 with \Eref{loss1}.
Regarding the dataset aspect, using LaughTalk$_{\text{CELEB}}$ yields significant improvement in both LVE and EFD metrics. 
This indicates that the data from expressive in-the-wild videos helps the model learn to animate both talking and laughing.
While utilizing the complete LaughTalk dataset could potentially yield even better performance, training with all the data at once might be complex due to the intertwined verbal and non-verbal signals.
Thus, as proposed, a two-stage training framework addresses these challenges and yields the best overall results in the evaluated metrics.
\section{Applications}
Unlike existing methods that utilize mesh vertex representations for 3D talking head models~\cite{voca,karras2017audio,codetalker,faceformer, meshtalk}, our approach employs FLAME parameters to represent facial motion. 
While the prior arts are limited to a fixed number of pre-defined identity templates for 3D talking heads, the use of FLAME parameters in our method offers more distinctive and user-friendly control. 
Moreover, these parameters often serve as control signals for rigging neural avatars. 
This opens up interesting and practical applications as follows.

\paragraph{User controllable identities}
\begin{figure}[tp]
    \centering
    \includegraphics[width=1\linewidth]{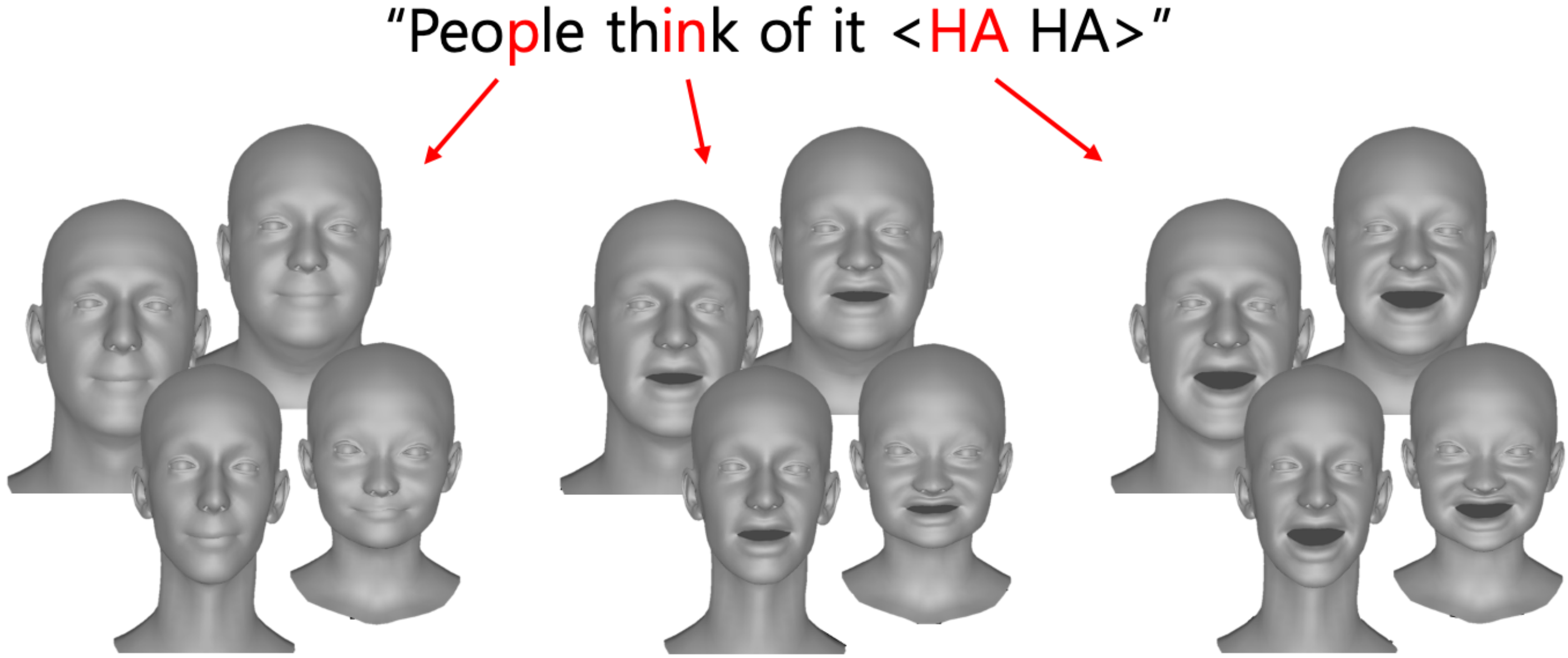}
    \caption{\textbf{User controllable identities.} As we use the FLAME parameters for facial representations, our model can synthesize 3D talking heads with diverse identities by changing the $\mathbf{\beta}$ parameter.}
    \label{fig:app1}
\end{figure}

Given the flexibility of our model in defining the shape parameter $\boldsymbol{\beta}$,
we can easily manipulate or even input $\boldsymbol{\beta}$ extracted from a reference image.
As shown in \Fref{fig:app1}, with the same input speech, our model can synthesize 3D talking heads with diverse identities.
Combined with the 3D reconstruction methods~\cite{mica, deca}, we can finely control the identity of the 3D talking head by users.

\begin{figure}[tp]
    \centering
    \includegraphics[width=1\linewidth]{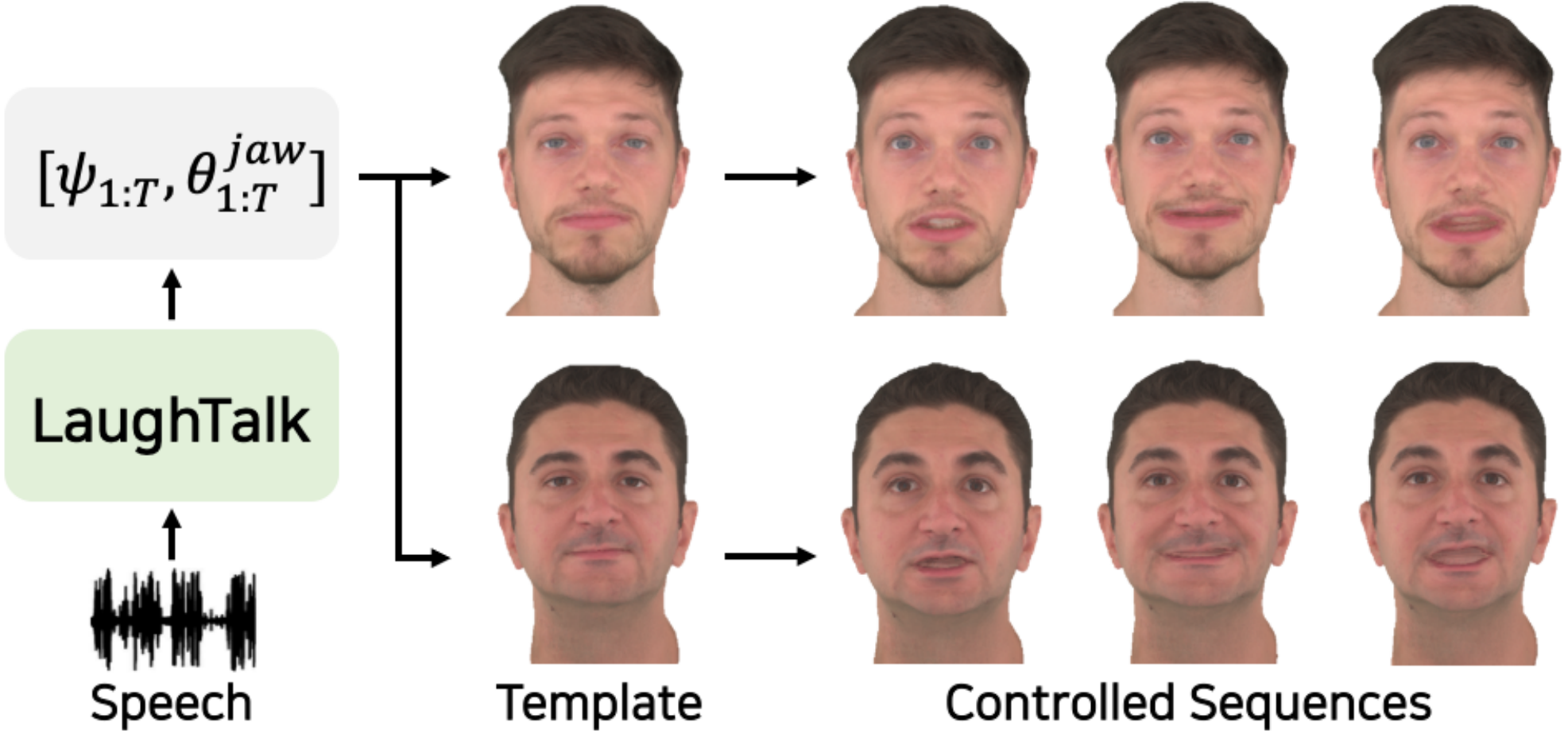}
    \vspace{-6mm}
    \caption{\textbf{Rigging neural avatars.} Our model can be used for rigging the photorealistic neural avatar from speech. Here, $\boldsymbol{\psi}_{1:T}$ and $\boldsymbol{\theta}^{\text{jaw}}_{1:T}$ denote the sequence of expression and jaw pose parameters predicted by our model from input speech. These predicted parameters serve as control signals for NHA~\cite{grassal2022neural}.}
    \vspace{-2mm}
    \label{fig:app2}
\end{figure}

\paragraph{Rigging neural avatars}
In addition to synthesizing 3D talking meshes, our model also offers FLAME parameters for avatar rigging. Recent studies have shown that the FLAME parameters can effectively serve as control signals for rigging photorealistic neural avatars~\cite{insta, grassal2022neural}. Figure~\ref{fig:app2} showcases several examples produced by combining our approach with NHA~\cite{grassal2022neural}. The expression and jaw pose parameters that our model predicts are directly fed into NHA to control the avatar. The results clearly illustrate that our method serves as a conduit for rigging photorealistic avatars with diverse speech inputs.

\section{Conclusion}
In this work, we introduce a novel task of animating a speech-driven 3D face, capable of expressing both verbal and non-verbal signals.
We especially focus on laughter as a non-verbal signal, given its importance in social interactions.
To approach this task, we curate the LaughTalk dataset, consisting of diverse in-the-wild 2D videos paired with pseudo ground truth 3D data.
Furthermore, we propose LaughTalk, a two-stage training baseline that can synthesize lip articulation and facial motion, synchronized to the input speech and laughter.
Our extensive experiments show the efficacy of our approach, demonstrating strong performance in lip and laughter synchronization with speech, and evoking a sense of intimacy by accurately reflecting the laughter signal in the 3D talking head model. 
We would like to note that our proposed design choice and learning approach are independent of the specific non-verbal signal.
As we primarily tackle the prominent non-verbal signal of laughter, future research could explore extending our model to convey other essential non-verbal cues, such as crying or shouting, thereby broadening its application scope.

{\small
\bibliographystyle{ieee_fullname}
\bibliography{egbib}
}

\clearpage
\appendix
\renewcommand{\thefigure}{S\arabic{figure}}
\setcounter{figure}{0}
\noindent {\LARGE \textbf{Appendix}}
\vspace{8mm}

The contents in this supplementary material are as follows: A. Training details for counterpart methods (\Sref{secA}), B. Details of the emotional feature distance (\Sref{secB}), and C. Details of the user study (\Sref{secC}). 
We recommend viewing the supplementary video, which showcases generated 3D face animations from speech and laughter.
\begin{figure*}[th]
  \centering
\includegraphics[width=1\linewidth]{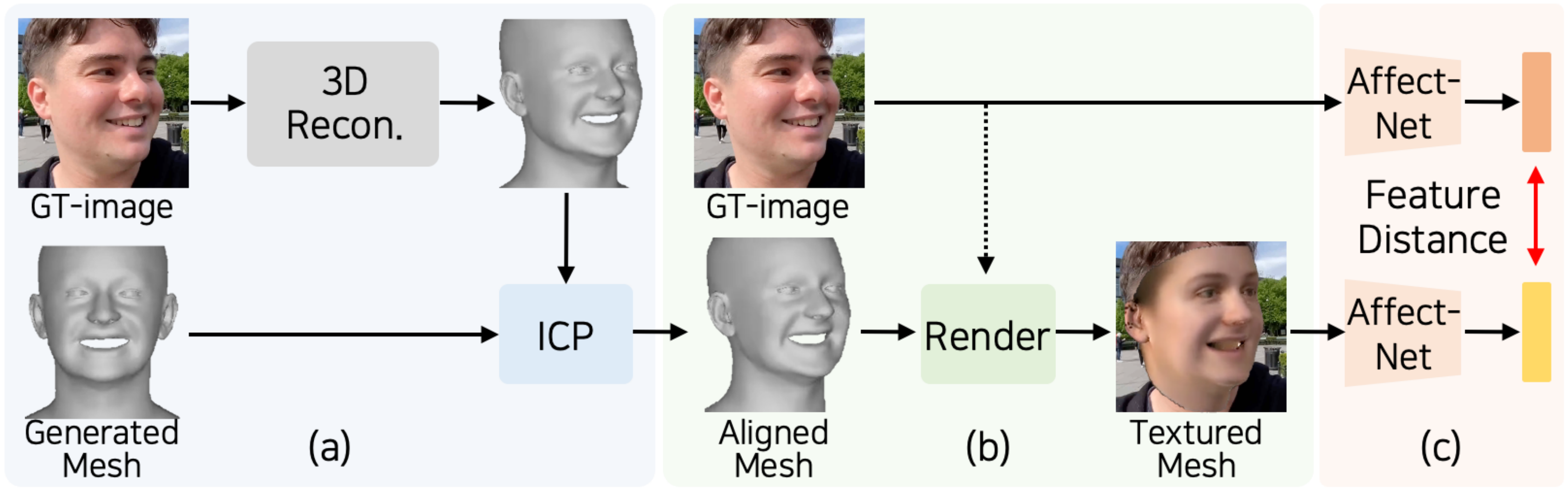}
   \caption{\textbf{Measuring emotional feature distance.} We present the Emotional Feature Distance (EFD), a perceptual metric designed to evaluate the synchronization of facial movements with laughter. 
   To calculate this metric, first, (a) we align the speech-driven generated mesh vertices with the mesh reconstructed from the ground truth image of the original video. 
   Next, (b) we render the generated mesh and texturize it using the texture map extracted from the ground-truth image. 
   Finally, (c) we feed both the ground truth image and the textured mesh into AffectNet~\cite{mollahosseini2017affectnet} and measure the $\ell_2$ feature distance.}
   \label{fig:efd}
\end{figure*}
\begin{figure*}[tp]
  \centering
\includegraphics[width=1\linewidth]{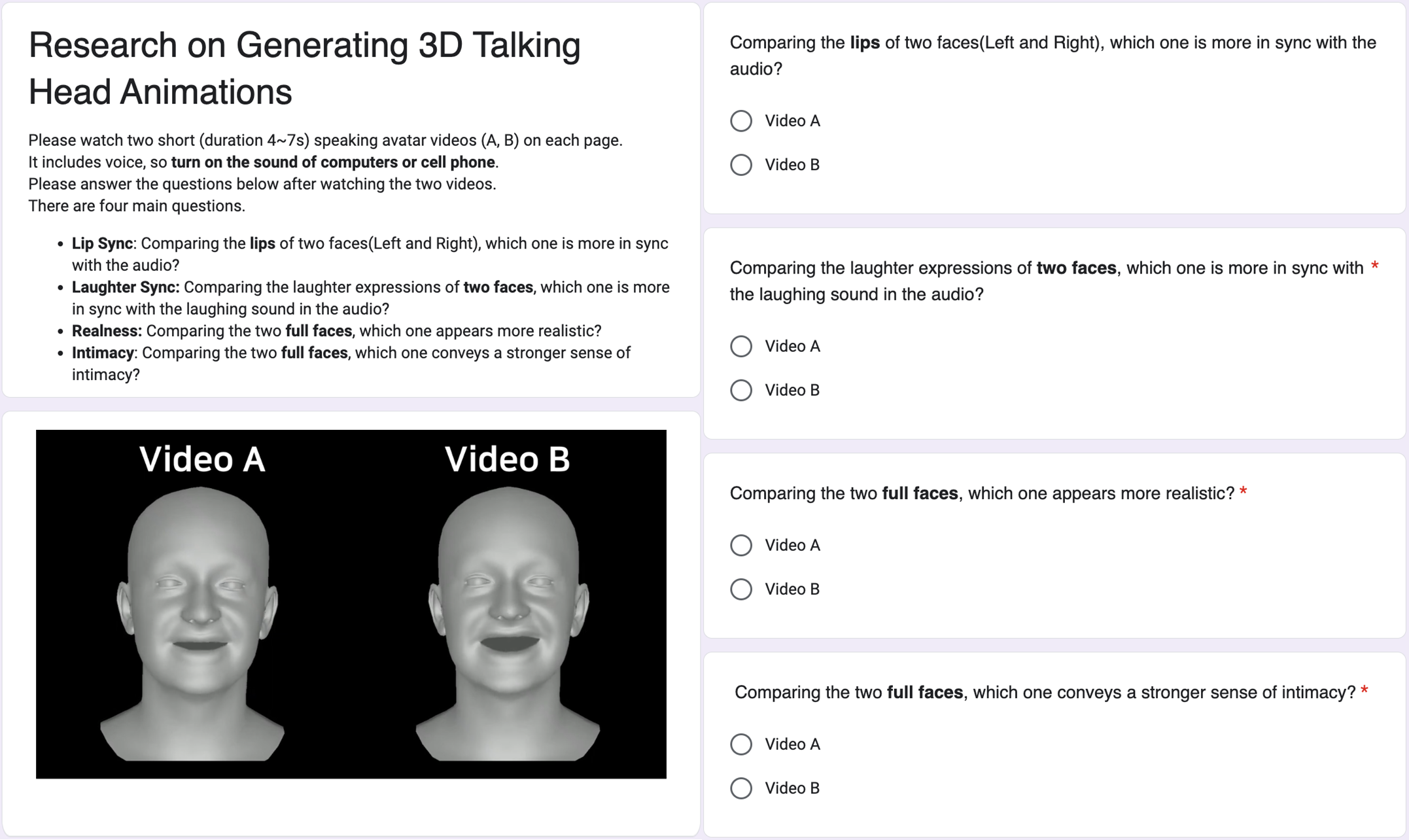}
\caption{\textbf{Example of a user study experiments.} Each page contains a pair of generated 3D talking head videos for comparative analysis accompanied by four questions designed to assess the performance of our model.}
   \label{fig:user_s1}
\end{figure*}
\section{Training details for counterpart methods}\label{secA}
As the pre-trained models of existing methods are not trained to capture the laughing expression, we retrain the existing methods~\cite{faceformer, codetalker, voca} with our proposed LaughTalk dataset to ensure fair qualitative and quantitative comparison. Here, we provide the training details of the existing methods. 

\paragraph{CodeTalker}
CodeTalker~\cite{codetalker} comprises two stages: the first stage involves learning generic motion through discrete tokens of a codebook, and the second stage focuses on generating 3D talking heads using these discrete tokens. 
Initially, we attempted to train only the second stage model with the LaughTalk dataset, while utilizing the pre-trained first stage model. 
However, due to the first stage model's lack of exposure to diverse and expressive talking heads, the model still failed to generate laughing expressions in the second stage. 
Therefore, we trained the first stage model on our dataset, encompassing diverse laughing expressions, and subsequently trained the second stage model with the same dataset. 
We followed the training scheme of the official code\footnote{\url{https://github.com/Doubiiu/CodeTalker}.}.

\paragraph{FaceFormer and VOCA}
For FaceFormer~\cite{faceformer} and VOCA~\cite{voca}, we initiated pre-training using LaughTalk${_\text{MEAD}}$, which consists of neutral speech and corresponding 3D faces. 
Subsequently, fine-tuning was conducted on both models using LaughTalk${_\text{CELEB}}$, featuring laughing and speech data. 
Notably, attempts to train these models using the entire LaughTalk dataset resulted in mode-collapsed outputs. 
The training process adhered to the official code of each method\footnote{\url{https://github.com/EvelynFan/FaceFormer}.\\\url{https://github.com/TimoBolkart/voca}.}.

\section{Details of the emotional feature distance}\label{secB}
As discussed in the main paper, relying solely on measuring the lip vertex error (LVE) is insufficient to accurately assess facial movement synchronization to laughter. To address this limitation, we introduce Emotional Feature Distance (EFD) as a perceptual metric for evaluating laughter synchronization (\Fref{fig:efd}). To compute the EFD, we utilize an off-the-shelf emotion recognition model, AffectNet~\cite{mollahosseini2017affectnet}. 
Using this model, we measure the average feature distance between sequences of images rendered from the generated mesh vertices and the image frames sourced from the ground truth 2D video.

However, one challenge to note is the fixed pose inherent in existing 3D talking head generation methods (ours included), especially when compared to the diverse head movements present in the ground truth video frames.
This discrepancy in head movement may result in misalignment between the generated meshes and the corresponding ground truth images, leading to less meaningful metric evaluations.

To mitigate this issue, we employ the iterative closest point (ICP) algorithm to align the generated meshes with the ground truth images (\Fref{fig:efd} (a)). Specifically, we begin by reconstructing a face mesh for each ground truth image using EMOCA~\cite{emoca}. The ICP algorithm then computes the rigid transformation matrix between the generated mesh and the mesh reconstructed from the ground truth images. This process is facilitated by the known correspondence between the vertex indices of the two meshes. 
The rigid transformation is subsequently applied to the generated mesh vertices, aligning them with the mesh of the ground truth image. 
We then proceed to texturize the aligned mesh with the texture map of the ground truth image and overlay it on top of the ground truth image (\Fref{fig:efd} (b)). 
Lastly, we feed both the rendered meshes and the ground truth images to the AffectNet and measure the $l_2$ distance between the extracted features, thus obtaining the EFD (\Fref{fig:efd} (c)).

\section{Details of the user study}\label{secC}
We conduct a user study to assess the performance of our method compared to the existing methods from a human perception standpoint. 
Our user study questionnaire interface is illustrated in \Fref{fig:user_s1}. 
During the study, participants watched two generated 3D talking head videos and responded to four questions, without any time constraints. 
The user study comprises a total of 15 sets, each consisting of 2 videos and featuring four questions in each set.
Our study includes 50 participants, encompassing individuals both within and outside the research field. 
The questions we ask to the participants are as follows:
\begin{itemize}
    \item Lip Sync: Comparing the lips of two faces (Left and Right), which one is more in sync with the audio?
    \item Laughter Sync: Comparing the laughter expressions of two faces, which one is more in sync with the laughing sound in the audio?
    \item Realness: Comparing the two full faces, which one appears more realistic?
    \item Intimacy: Comparing the two full faces, which one conveys a stronger sense of intimacy?
\end{itemize}
\end{document}